\documentclass[runningheads]{llncs}

\usepackage{eccv}

\usepackage{eccvabbrv}

\usepackage{graphicx}
\usepackage{booktabs}

\usepackage[accsupp]{axessibility}  

\usepackage{caption,subcaption,multirow,hhline,makecell,graphicx}
\usepackage{hyperref}

\usepackage{orcidlink}

\begin{document}

\title{PoseAugment: Generative Human Pose Data Augmentation with Physical Plausibility
    for IMU-based Motion Capture}

\titlerunning{PoseAugment}

\author{Zhuojun Li\inst{1,2}\thanks{Email: lizj23@mails.tsinghua.edu.cn}\orcidlink{0000-0003-4374-9452} \and
Chun Yu\inst{1,2,3}\thanks{Corresponding author: chunyu@mail.tsinghua.edu.cn}\orcidlink{0000-0003-2591-7993} \and
Chen Liang\inst{1,2}\orcidlink{0000-0003-0579-2716} \and
Yuanchun Shi\inst{1,2,3}\orcidlink{0000-0003-2273-6927}}

\authorrunning{Li et al.}

\institute{Department of Computer Science and Technology, Tsinghua University, China \and
Key Laboratory of Pervasive Computing, Ministry of Education, China \and
Qinghai University, China}

\maketitle

\begin{abstract}
    The data scarcity problem is a crucial factor that hampers
    the model performance of IMU-based human motion capture.
    However, effective data augmentation for IMU-based motion capture is challenging,
    since it has to capture the physical relations and constraints of the human body,
    while maintaining the data distribution and quality.
    We propose PoseAugment, a novel pipeline incorporating VAE-based
    pose generation and physical optimization.
    Given a pose sequence, the VAE module generates infinite poses with both high fidelity
    and diversity, while keeping the data distribution.
    The physical module optimizes poses to satisfy physical constraints
    with minimal motion restrictions.
    High-quality IMU data are then synthesized from the augmented poses
    for training motion capture models.
    Experiments show that PoseAugment outperforms previous data augmentation
    and pose generation methods in terms of motion capture accuracy,
    revealing a strong potential of our method to alleviate the data collection
    burden for IMU-based motion capture and related tasks driven by human poses.
\end{abstract}

\section{Introduction}\label{sec:introduction}

\begin{figure*}[ht]
    \centering
    \includegraphics[width=\linewidth]{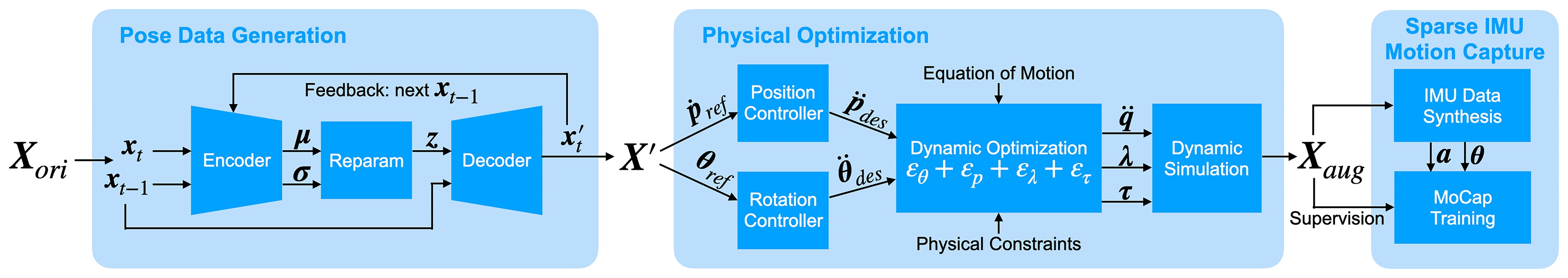}
    \caption{Method overview.
        Given an original pose sequence
        $\boldsymbol{X}_{ori}=\{\boldsymbol{x}_1,\dots\boldsymbol{x}_T\}$,
        we learn a VAE model to generate new poses $\boldsymbol{X}'$
        frame-by-frame autoregressively.
        It captures the motion variance and can generate infinite poses
        within this distribution.
        Then, the motion jitter and artifacts are optimized by solving a
        quadratic optimization problem, which is based on a dual position ($\boldsymbol{p}$)
        and rotation ($\boldsymbol{\theta}$) PD controller and physical constraints
        on reaction forces $\boldsymbol{\lambda}$ and torques $\boldsymbol{\tau}$.
        The final augmented poses $\boldsymbol{X}_{aug}$ can be used to augment the dataset by synthesizing IMU data.}
    \label{fig:overview}
\end{figure*}

Human motion capture (MoCap) with IMUs has become a rising topic in recent research
\cite{Von17,Huang18,Yi21,Yi22,Mollyn23,Jiang22,Tom23},
due to its advantages in terms of power efficiency, privacy preservation, and usability
compared with CV-based methods.
However, training MoCap models requires large IMU and corresponding pose data,
which are normally collected via professional MoCap systems,
such as OptiTrack \cite{OptiTrack} and Vicon \cite{Vicon},
which are both expensive and time-consuming.
Xsens\cite{Xsens} reduces the cost by using 17 IMUs, but it is still inconvenient for individuals.
Besides, human motion distribution is also highly personalized and task-related.
The IMU signals can also be affected by different mount positions
and hardware errors, resulting in low data transferability across different tasks.

To avoid the data collection burden, we need effective data augmentation methods.
Current methods synthesize IMU data directly from open-source
pose datasets\cite{Yi21,Yi22,Jiang22,Tom23}, such as AMASS \cite{Mahmood19}.
However, it is unsuitable for specialized tasks where the training samples are hard to collect
and have a unique distribution, like in disease/sports analysis \cite{Das11,Supej10}.
Researchers also leveraged noise-based methods to directly augment the IMU data\cite{Yi21, Yi22},
but these methods are unable to capture the physical constraints among multiple IMU nodes,
thus only having marginal improvements.
Recently, generative deep models (\eg VAE, diffusion) demonstrated a huge potential
in generating natural human motions for animations.
However, these methods mainly use high-level texts or actions to condition pose generation,
which will affect the data distribution and quality of the original dataset.
We found generating poses that are close to the original data is more suitable
for IMU-based motion capture in our evaluations.

We propose PoseAugment, a two-stage human pose data augmentation pipeline for
IMU-based motion capture (\cref{fig:overview}).
Given a reference pose, the first stage leverages a VAE \cite{Oord17}
model to generate infinite new poses, covering the original data space comprehensively.
The encoder encodes the difference between adjacent frames into a latent representation.
Then, the decoder reconstructs the next frame based on the previous frame
and the latent vector.
The generation process is guided by the original data frame-by-frame autoregressively,
which maintains the low-level motion distribution with the original data.

The second stage optimizes poses with physical constraints to improve
motion naturalness, as the generated poses may have artifacts (jitter, unnatural motion).
We apply a dual PD controller \cite{Yi22} to provide positional
and rotational motion references.
Different from prior work \cite{Shimada20,Yi22}, our method only has minimal
physical constraints, which do not assume the motion to be on flat ground and
do not need contact labeling, thus supporting a wider range of motions
(\eg climbing stairs, skating) and achieving a soft estimation of reaction forces.

After that, high-quality IMU data could be synthesized from the augmented poses for MoCap training.
We compared our data augmentation method with adding random noise,
and other generative methods (MotionAug\cite{Maeda22}, ACTOR\cite{Petrovich21}, MDM\cite{Tevet22})
on training MoCap models (TransPose\cite{Yi21}).
Results show that PoseAugment outperformed these methods in pose accuracy.
Future works can leverage PoseAugment as a pre-trained generative model to alleviate
the data collection burden in IMU-based MoCap and other tasks driven by human poses.

Our main contribution in this paper is two-fold:
\begin{enumerate}
    \item We propose PoseAugment, a novel pose data augmentation pipeline
        incorporating pose generation and physical optimization.
        Future works can benefit from our method to reduce the data collection burden.
    \item We thoroughly evaluate PoseAugment with previous pose generation methods,
        which shows a significant improvement in IMU-based motion capture.
\end{enumerate}

\section{Related Work}\label{sec:related_work}

We will first review the context of IMU-based motion capture,
then introduce IMU data augmentation, pose data generation, and physical optimization,
mainly from the perspective of their influences on the data quality.

\subsection{IMU-based Motion Capture}\label{ssec:imu_motion_capture}

To overcome the occlusion and privacy issues of CV-based motion capture methods,
a series of works focus on reconstructing human poses solely from sparse IMU sensors,
worn on key body joints. 

SIP \cite{Von17} first proposes a 6-IMU setting,
and regards the motion capture task as a non-convex optimization problem.
DIP \cite{Huang18} leverages bidirectional RNN to regress human poses,
providing a learning-based approach.
TransPose \cite{Yi21} proposes a three-stage network structure and achieves a
better regression accuracy both in pose measurements and global translations.
The follow-up study PIP \cite{Yi22} optimizes poses with physical constraints to fix motion artifacts,
but only supports motions on the ground.
TIP\cite{Jiang22} and DiffIP\cite{Tom23} applied Transformer and diffusion models to this task,
increasing the model robustness to different IMU configurations.
IMU Poser \cite{Mollyn23} further reduces IMU numbers to daily settings (1 to 3),
making inertial motion capture more pervasive.
AvatarPoser\cite{Jiang22b}, BoDiffusion\cite{Castillo23}, and AGRoL\cite{Du23}
aim to reconstruct full poses only from positions and rotations of VR devices,
which could also be regarded as preprocessed IMU data.

All of these works require large IMU and pose data for model training.
Apart from network improvements, our motivation is that augmenting data with high quality
is equally crucial to model performance, especially for tasks with little available data to use.
Therefore, human pose data augmentation in the context of IMU-based motion
capture is worthy of exploring.

\subsection{IMU Data Augmentation}

Inspired by image augmentation in CV fields, numerous methods have been
adapted to IMU signals, including noise-based methods (\eg Jitter)\cite{Wen21,Iwana21,Xu22}
and generative methods (\eg GAN)\cite{Chen21}.
However, these methods only work in classification tasks,
since the data labels need not change after data augmentation.
But for IMU-based MoCap, the semantic changes in IMU signals 
after augmentation are ambiguous, making it hard to augment poses accordingly.

Due to this restriction, only TransPose \cite{Yi21} among works in
\cref{ssec:imu_motion_capture} leverages Jitter to augment data.
We demonstrate that direct augmentation on IMU signals with traditional
methods only has marginal improvements to MoCap models (\cref{ssec:evaluation}).
Our method, on the other hand, augments poses and synthesizes
the corresponding IMU signals, achieving better semantic control.

\subsection{Pose Data Generation}\label{ssec:pose_data_generation}

Generating natural human poses is a basic problem in computer vision.
From the model perspective, some works use autoencoders to reconstruct human poses,
like VAE\cite{Guo20,Petrovich21,Maeda22,Ling20,Won22,Peng22,Tessler23}.
Recent works also leverage diffusion models to generate human poses
\cite{Guo22,Zhang24,Tevet22,Chen23,Yuan23,Karunratanakul23}, inspired by the success of AIGC.
From the perspective of generation goals, some works generate similar poses
based on a reference pose (M2M)\cite{Maeda22,Henter20,Ling20,Gong21,Rogez16},
while a large amount of research converts data from other modalities to poses,
like action-to-motion (A2M)\cite{Guo20,Petrovich21}
and text-to-motion(T2M) \cite{Guo22,Zhang24,Tevet22,Chen23,Yuan23,Karunratanakul23}.

To augment poses for MoCap training, our goal is to generate more pose samples while
maintaining the data distribution.
To achieve the fine-grained motion control, we choose to augment poses in the M2M manner,
instead of A2M and T2M, where the motions can not be controlled precisely.
MotionAug\cite{Maeda22} proposed a similar VAE/IK and optimization solution like ours.
But it needs users to annotate IK keyframes and is only limited to 8 motion types.
So, we do not regard it as an off-the-shelf general-purpose solution.
MVAE\cite{Ling20} proposed a frame-to-frame generation design.
A new frame is predicted using the previous frame and the difference between frames.
This is more suitable for IMU data synthesis since the IMU signals
are also frame-based when taking differentials on poses.
Therefore, we adopted this design and improved the network structure
for our task (\cref{ssec:pose_data_generation}).

\subsection{Physical Optimization}

Physics-based optimization is used to correct motion artifacts after motion capture.
Some works use direct constraints (\eg foot contacts)
to restrict body motions \cite{Du19,Zou20,Zanfir18}.
Some works impose physical constraints as loss terms when training
motion capture models \cite{Shi20,Habermann19,Mehta17}.
Recent optimization-based methods estimate the physical properties of the
human body (\eg torques, reaction forces).
They consider the human body as an articulated rigid body system,
and calculate the dynamic properties based on the equation of motion
\cite{Li19,Davis20,Wei10,Zell17,Liu10}.
Then, the rigid body system could be actuated by the optimized
dynamic properties to generate motions \cite{Zheng13,Shimada20,Yi22}.

PIP \cite{Yi22} proposes a dual PD controller to calculate desired motion accelerations.
However, it only supports motions on the ground and needs contact point labeling.
We adopted the dual PD controller with a looser assumption and no contact information,
supporting a wider variety of motions (\eg climbing stairs).

\section{Method}\label{sec:method}

The goal of our method is to augment human pose data for MoCap training.
The input is a pose sequence, and the output is augmented poses
and the corresponding IMU signals (including accelerations and rotations at key body joints).
PoseAugment incorporates two stages:
(1) \textbf{Pose Data Generation} generates pose sequences with both
high fidelity and diversity (\cref{ssec:pose_data_generation});
(2) \textbf{Physical Optimization} corrects motion artifacts
to improve motion naturalness (\cref{ssec:physical_optimization}).
Then, the IMU data are synthesized from poses by taking derivatives.

\subsection{Pose Data Generation}\label{ssec:pose_data_generation}

We aim to generate new poses with minimal reconstruction errors
while diverse enough to cover the motion space, which is a typical trade-off problem.
All pose data are in SMPL\cite{Loper15} format, which considers the
human body as a 24-joint articulated rigid body system.
We will first introduce the data representation,
and then detail the VAE model structure, training, and inferencing.

\subsubsection{Motion Frame Representation.}\label{sssec:motion_frame_representation}

In this stage, we represent pose data as a motion frame sequence
$\boldsymbol{X} = [\boldsymbol{x}_1, \boldsymbol{x}_2, \dots, \boldsymbol{x}_T]$,
each frame
\begin{equation}\label{eq:motion_frame}
    \boldsymbol{x} = [\boldsymbol{p}_{root}, \boldsymbol{v}_{root},
        \boldsymbol{\theta}_{root}, \boldsymbol{p}_{joint},
        \boldsymbol{v}_{joint}, \boldsymbol{\theta}_{joint}] \in \mathbb{R}^{240}
\end{equation}
where $\boldsymbol{p}, \boldsymbol{v}, \boldsymbol{\theta}$ represent 3D position,
3D velocity and 6D rotation \cite{Zhou19} respectively.
The subscript \textit{root} means the SMPL root joint (pelvis) in the global frame,
while \textit{joint} means other 19 SMPL joints (excluding L/R feet and L/R hands,
which are simplified as identical rotations) in the root frame.
So, the total dimension of a motion frame is $3+3+6+(3+3+6)\times19=240$.
To reconstruct poses from the augmented motion frames, we only use
the original $\boldsymbol{p}_{root}, \boldsymbol{\theta}_{root}$,
and the augmented $\boldsymbol{\theta}_{joint}$.
We added other $\boldsymbol{p}$ and $\boldsymbol{v}$ terms as an additional
reconstruction task for the VAE model to better learn global motion features.

\subsubsection{VAE Model Structure.}\label{sssec:vae_model_structure}

When designing the VAE model, the key rule is to make it powerful enough
to reconstruct accurate poses, while emphasizing the latent vector
to ensure pose diversity, thus covering the motion space comprehensively.
We leveraged several successful designs of MVAE \cite{Ling20},
including the autoregressive prediction and the MoE architecture\cite{Masoudnia14,Zhao24}.
We further propose several improvements to make the VAE model fit our goal better.

Given two frames $\boldsymbol{x}_{t-1}$ and $\boldsymbol{x}_{t}$,
the encoder first encodes them into latent mean and standard deviation vectors
$\boldsymbol{\mu}, \boldsymbol{\sigma} \in \mathbb{R}^{40}$.
After reparameterization \cite{Oord17}, the latent vector
$\boldsymbol{z}\in\mathbb{R}^{40}$, together with frame $\boldsymbol{x}_{t-1}$,
will be decoded to predict frame $\boldsymbol{x}_{t}'$,
which will be used as the next $\boldsymbol{x}_{t-1}$ to generate poses
autoregressively.

Unlike MVAE, we incorporate two separate residual blocks in the encoder
and one residual block in the decoder.
They deepen the network while also ensuring that the motion and latent features
would not be ignored during inferencing, which ensures both fidelity and diversity.
Furthermore, we add a separate layer to expand the latent vector to
$\boldsymbol{z}_{exp}\in\mathbb{R}^{240}$ (with the same length as a motion frame)
to extract the compressed latent features.
Then, $\boldsymbol{z}_{exp}$ is input to both the MoE gate network and decoding
layers, together with $\boldsymbol{x}_{t-1}$.
The VAE structure details are shown in Appendix A.1.

\subsubsection{VAE Model Training.}\label{sssec:vae_model_training}
Following the former practice\cite{Yi21,Yi22}, we used AMASS\cite{Mahmood19}
and DIP train split\cite{Huang18} to train the VAE model,
and evaluated it on the DIP test split.
The training dataset contains about 45.6h poses sampled at 60Hz.
 
To make pose prediction stable, we trained the VAE model with mini-batches
and scheduled sampling, following MVAE\cite{Ling20}.
The pose sequences are first cut into mini-batches with length $L$.
Within each mini-batch, each pose prediction $\boldsymbol{x}_t'$ will be used
as the next conditione frame $\boldsymbol{x}_{t-1}$ with probability $(1-p)$,
(ground truth $\boldsymbol{x}_{t}$ with probability $p$).
$p$ will decrease gradually from $1$ to $0$ during training,
transforming the training process from fully supervised to fully autoregressive.
This method will make the VAE model robust to self-prediction errors.

In the $\beta$-VAE design\cite{Higgins16}, the training loss is
$loss_{reconst} + \beta \cdot loss_{KL}$, where $\beta$ is introduced
to balance the reconstruction accuracy and the latent vector diversity.
We found the pose diversity is highly sensitive to $\beta$.
As a result, we fine-tuned $\beta$ to be $3\times10^{-3}$ empirically,
which achieved a balance of pose fidelity and diversity (\cref{ssec:fidelity_and_diversity}).
More training details are provided in Appendix A.2.

\subsubsection{VAE Model Inferencing.}\label{sssec:vae_model_inferencing}

After training, the VAE model is used to augment poses autoregressively,
during which the prediction errors may accumulate in motion frames,
resulting in unrecoverable deviations from the ground truth.
Since we need to augment poses with lengths far longer than the mini-batch,
(200 frames for\cite{Yi21,Yi22}), how to ensure the temporal stability
is a crucial problem.
The general idea is to let the ground truth guide pose generation.
We designed two techniques named \textbf{best sampling}
and \textbf{motion refinement} to tackle it.

For \textit{best sampling}, at each timestamp, we use $\boldsymbol{x}_{t-1}$ and
$\boldsymbol{x}_t$ to repeatedly generate $N$ different predictions
$\{\boldsymbol{x}_{t}'^{(1)}, \boldsymbol{x}_{t}'^{(2)}, \dots, \boldsymbol{x}_{t}'^{(N)}\}$,
from which we choose the closest frame to $\boldsymbol{x}_{t}$ in MSE error as the final prediction.
It can discard predictions with large errors to make data augmentation more stable,
but may harm pose diversity on the other hand.
We choose $N=2$ in our evaluations.

Furthermore, \textit{motion refinement} restricts frame values in a reasonable range,
to avoid error accumulation.
For $\boldsymbol{p}_{root}$ and $\boldsymbol{p}_{joint}$, we let their distances to
$\boldsymbol{p}_{GT}$ (ground truth) within $d_{p}=15cm$.
For $\boldsymbol{v}_{root}$ and $\boldsymbol{v}_{joint}$, we force their values
to be no larger than $d_{v}\cdot\boldsymbol{v}_{GT}$ and no smaller than
$(1/d_{v})\cdot\boldsymbol{v}_{GT}$, where $d_{v}=2.0$.
For $\boldsymbol{\theta}_{root}$ and $\boldsymbol{\theta}_{joint}$,
since each 6D rotation consists of two orthogonal unit vectors
(but networks do not ensure that), we renormalized them after each prediction.

They significantly improve the prediction stability.
Tuning these hyperparameters also helps to balance the pose fidelity and diversity after training.

\subsection{Physical Optimization}\label{ssec:physical_optimization}

The pose data augmented by the VAE model are purely kinematics-based, but do not consider
how the human body is actuated by dynamic properties like forces and torques.
In this stage, we perform physical optimization on pose data to
improve motion naturalness and temporal consistency.
We regard the human body as a 24-joint articulated rigid body system
as introduced by\cite{Featherstone14}, which is actuated by the
internal torques and external reaction forces.
Then, the optimal dynamic properties are calculated by solving
a quadratic optimization problem, which satisfies the motion of equation\cite{Featherstone14},
the desired accelerations given by a dual PD controller\cite{Yi22},
and physical constraints on reaction forces and torques.
Using these optimized dynamic properties, the final poses could be simulated.

\subsubsection{Problem Definition.}\label{sssec:problem_definition}

Similar to PIP\cite{Yi22}, we use a floating-base rigid body system as the human body,
which is controlled by reaction forces, torques,
and non-linear effects like gravity and Coriolis forces.
Pose data in this stage are represented in RBDL format\cite{Felis16},
which has the same $J=24$ joints as SMPL\cite{Loper15} with a different order
(root indices are both $0$).
We denote joint positions as $\boldsymbol{p}\in\mathbb{R}^{3J}$
and joint rotations as $\boldsymbol{\theta}\in\mathbb{R}^{3J}$ (in Euler angles).
The corresponding velocities and accelerations would be
$\dot{\boldsymbol{p}},\ddot{\boldsymbol{p}},\dot{\boldsymbol{\theta}},
\ddot{\boldsymbol{\theta}}$ respectively.
A motion frame is represented by the root position and joint rotations
(in parent frames) as $\boldsymbol{q}=[\boldsymbol{p}_{root},
\boldsymbol{\theta}]\in\mathbb{R}^{N}$, with $N=3+3J$.
The system is actuated by internal torques $\boldsymbol{\tau}\in\mathbb{R}^{N}$
on each DoF and external reaction forces $\boldsymbol{\lambda}\in\mathbb{R}^{3(J-1)}$.

The system follows the equation of motion\cite{Featherstone14}
\begin{equation}\label{eq:equation_of_motion}
    \boldsymbol{\tau} + \boldsymbol{J}(\boldsymbol{q})^T\boldsymbol{\lambda}
        = \boldsymbol{M}(\boldsymbol{q})\ddot{\boldsymbol{q}}
        + \boldsymbol{h}(\boldsymbol{q},\ddot{\boldsymbol{q}})
\end{equation}
where $\boldsymbol{M}\in\mathbb{R}^{N\times N}$ is the inertial matrix;
$\boldsymbol{h}\in\mathbb{R}^{N}$ is the non-linear effect term; 
$\boldsymbol{J}\in\mathbb{R}^{3(J-1)\times N}$ is the
Jacobians of $(J-1)$ joints \cite{Featherstone14}.
Each joint Jacobian $\boldsymbol{J}_i\in\mathbb{R}^{3\times N}$ converts
the motion frame velocity to the global joint velocity as
\begin{equation}\label{eq:jacobian}
    \dot{\boldsymbol{p}}_i = \boldsymbol{J}\dot{\boldsymbol{q}}
\end{equation}

Derivations can be found from rigid body dynamics\cite{Featherstone14}.
Unlike PIP\cite{Yi22}, we do not assume $\boldsymbol{\lambda}$ all come from flat ground.
The reaction forces can act on any non-root joint to support a wider range of motions,
which we will detail later.

\subsubsection{Dynamic Optimization.}\label{sssec:dynamic_optimization}

The optimization problem can be formalized as:
\begin{equation}\label{eq:optimization}
    \begin{aligned}
        \mathop{\arg\min}_{\ddot{\boldsymbol{q}},\boldsymbol{\lambda},\boldsymbol{\tau}}
            \quad & (\varepsilon_{\theta} + \varepsilon_{p}
            + \varepsilon_{\lambda} + \varepsilon_{\tau})
            & \rm{(energy\ terms)} \\
        s.t.\quad & \boldsymbol{\tau} + \boldsymbol{J}^T\boldsymbol{\lambda}
            = \boldsymbol{M}\ddot{\boldsymbol{q}} + \boldsymbol{h}
            & \rm{(equation\ of\ motion)} \\
        \quad & |\dot{\boldsymbol{p}}^T\boldsymbol{\lambda}|\le\delta
            & \rm{(stationary\ support)} \\
        \quad & \boldsymbol{\lambda}\in\mathcal{F}
            & \rm{(friction\ constraint)} \\
    \end{aligned}
\end{equation}
This optimization is a Quadratic Programming problem, which gives the best
$\ddot{\boldsymbol{q}},\boldsymbol{\lambda},\boldsymbol{\tau}$
that minimize the four energy terms and satisfy the equation
of motion and physical constraints on reaction forces.
The optimization problem is solved using the Augmented Lagrangian
algorithm\cite{Bambade22} with sparse matrices.

In this optimization problem, the dual PD controller terms
$\varepsilon_{\theta}$ and $\varepsilon_{p}$ and the friction constraint
$\boldsymbol{\lambda}\in\mathcal{F}$ are similar to PIP\cite{Yi22}.
Different from PIP, we have a more general assumption on the regularization
and constraint of the reaction forces $\boldsymbol{\lambda}$,
which supports a wider range of motions and does not need contact labeling.
The energy terms and physical constraints are detailed as follows.

\paragraph{Dual PD Controller Terms $\varepsilon_{\boldsymbol{\theta}}$
and $\varepsilon_{\boldsymbol{p}}$.}
Given a reference motion, the dual PD controller is used to calculate
the desired positional and rotational accelerations
$\ddot{\boldsymbol{p}}_{des}$ and $\ddot{\boldsymbol{\theta}}_{des}$.
To achieve this, we first need to calculate the reference joint velocities
$\dot{\boldsymbol{p}}_{ref}$ and rotations $\boldsymbol{\theta}_{ref}$
of the next timestamp.
Given the current motion status of the optimized system,
$\ddot{\boldsymbol{\theta}}_{des}$ can be written as
\begin{equation}\label{eq:rotation_controller}
    \ddot{\boldsymbol{\theta}}_{des}
        = k_{p_{\theta}}(\boldsymbol{\theta}_{ref}-\boldsymbol{\theta}_{cur})
        - k_{d_{\theta}}\dot{\boldsymbol{\theta}}_{cur}
\end{equation}
and $\ddot{\boldsymbol{p}}_{des}$ can be written as
\begin{equation}\label{eq:position_controller}
    \ddot{\boldsymbol{p}}_{des}
        = k_{p_{p}}(\dot{\boldsymbol{p}}_{ref}\Delta t)
        - k_{d_{p}}\dot{\boldsymbol{p}}_{cur}
\end{equation}
where $\Delta t=(1/60)s$.
$k_{p_{\theta}}, k_{d_{\theta}}, k_{p_{p}}, k_{d_{p}}$ are the gain parameters,
which are empirically set to $1800,60,2400,60$
(related to how the motion kinematics are updated).

Then, the energy terms give the best $\ddot{\boldsymbol{q}}$ that minimizes
the distance of joint accelerations to
$\ddot{\boldsymbol{\theta}}_{des}$ and
$\ddot{\boldsymbol{p}}_{des}$:
\begin{equation}\label{eq:dual_pd_terms}
    \begin{aligned}
        & \varepsilon_{\theta} = ||\ddot{\boldsymbol{q}}_{3:}
            -\ddot{\boldsymbol{\theta}}_{des}||^{2} \\
        & \varepsilon_{p} = ||\ddot{\boldsymbol{p}}
            -\ddot{\boldsymbol{p}}_{des}||^{2},\quad
            \ddot{\boldsymbol{p}} = \dot{\boldsymbol{J}}\dot{\boldsymbol{q}}
            + \boldsymbol{J}\ddot{\boldsymbol{q}} \\
    \end{aligned}
\end{equation}
where $\ddot{\boldsymbol{p}}$ is the derivative of \cref{eq:jacobian}.
Readers can see \cite{Featherstone14,Yi22} for more details.

\paragraph{Regularization Terms $\varepsilon_{\boldsymbol{\lambda}}$
and $\varepsilon_{\boldsymbol{\tau}}$.}
$\boldsymbol{\lambda}$ represents the external reaction forces
on non-root body joints.
Previous works only consider ground reaction forces (GRF).
They assume the character moves on flat ground with no sliding.
To calculate GRF, these methods either need foot contact
prediction\cite{Shimada20,Shi20} or labeling with threshold-based methods\cite{Yi22}.
These approaches cannot deal with other common motions like
climbing stairs, holding the handrail with hands, sitting on a chair, \etc,
where the reaction forces may act on all body joints.

Instead, we propose a novel approach based on the \textbf{dis-to-root principle}.
We allow the reaction forces to act on any non-root joint to
support general human motions.
Noticing that the reaction forces largely come from the ground,
handrails, walls and act on body joints that are relatively far away from
the center of mass of the human body, we will give larger penalties
to joints closer to the root.
We first calculate the distance between root and joint $i$ as $d_{i}$,
then regularize $\boldsymbol{\lambda}$ as
\begin{equation}\label{eq:lambda_term}
    \varepsilon_{\lambda} = k_{\lambda}\mathop{\sum}_{i=1}^{23}
        ||\cfrac{1}{d_i}\boldsymbol{\lambda}_{i}||^{2}
\end{equation}
where $k_{\lambda}=0.02$ is the weight coefficient.
For motions on flat ground, the foot joints will have the largest reaction forces,
which is in accordance with previous work\cite{Shimada20,Shi20,Yi22}.
Note that this is only one requisite on $\boldsymbol{\lambda}$,
together with other constraints in \cref{eq:optimization}, we will make
the optimization for $\boldsymbol{\lambda}$ more reasonable.

For $\boldsymbol{\tau}$, the regularization term is written as
\begin{equation}\label{eq:tau_term}
    \varepsilon_{\tau} = k_{root}||\boldsymbol{\tau}_{:6}||^{2}
        + k_{joint}||\boldsymbol{\tau}_{6:}||^{2}
\end{equation}
where $k_{root}=0.05$ is the weight coefficient of the root joint
(the root position and rotation), while $k_{joint}=0.02$ is for
other joints \cite{Yi22}.

\paragraph{Stationary Support Constraint.}

To further restrict reaction forces, we propose the
\textbf{stationary support} constraint, which assumes all
the objects that provide reaction forces must be still.
Mathematically, it states
\begin{equation}\label{eq:statioanry_support}
    |\dot{\boldsymbol{p}}_i^T\boldsymbol{\lambda}_i| \le \delta,
    \quad (1 \le i \le 23)
\end{equation}
where $\dot{\boldsymbol{p}}_i$ and $\boldsymbol{\lambda}_i$
are the velocity of and the reaction force on joint $i$.
$\delta=10$ is a pre-defined threshold.

This equation has two connotations.
First, it prevents the joint velocity and the reaction force to be in the same direction,
which means the joint cannot move towards, or be pushed by the object
that exerts the reaction force.
Second, it allows the reaction force to be perpendicular to the velocity,
which is common in motions like skating or running on a treadmill.
This constraint cancels a large number of unreasonable reaction forces,
while still supporting any environment with stationary objects
(including but not limited to a ground plane).

\paragraph{Friction Constraint.} We assume the friction force on a joint cannot exceed
the maximum static friction force, while the joint can still slide on the object, as
\begin{equation}
    |\lambda_{x_i}|\le\mu\lambda_{y_i}, 
    |\lambda_{z_i}|\le\mu\lambda_{y_i},
    \quad (1 \le i \le 23)
\end{equation}
where $\mu=0.6$ is the friction coefficient, and $x,y,z$ denote the three dimensions.

\subsection{IMU Data Synthesis}

To train sparse IMU MoCap models, we need to synthesize the virtual IMU data,
including global accelerations and rotations
\footnote{Note that the raw IMU signals (in device local frames) would
first be processed into the global frame before motion capture in practice.}
at 6 body positions (2 wrists, 2 knees, head and pelvis).
Given a pose sequence $[\boldsymbol{p}_{root},\boldsymbol{\theta}_{root},
\boldsymbol{\theta}_{joint}]$, we first use Forward Kinematics\cite{Featherstone14}
to calculate the global joint and mesh positions of the human body.
Then, the accelerations and rotations on the 6 body positions could
be simulated by taking differentials.

\section{Experiments}\label{sec:experiment}

In this section, we will first show our qualitative and quantitative
evaluations on pose fidelity and diversity (\cref{ssec:fidelity_and_diversity}),
as well as on physical plausibility (\cref{ssec:physical_plausibility}).
Then, the data augmentation performance is tested on training MoCap models,
comparing with previous methods (\cref{ssec:evaluation}).

\subsection{Fidelity and Diversity}\label{ssec:fidelity_and_diversity}

\begin{figure}[ht]
    \centering
    \begin{subfigure}{0.24\linewidth}
        \includegraphics[width=\linewidth]{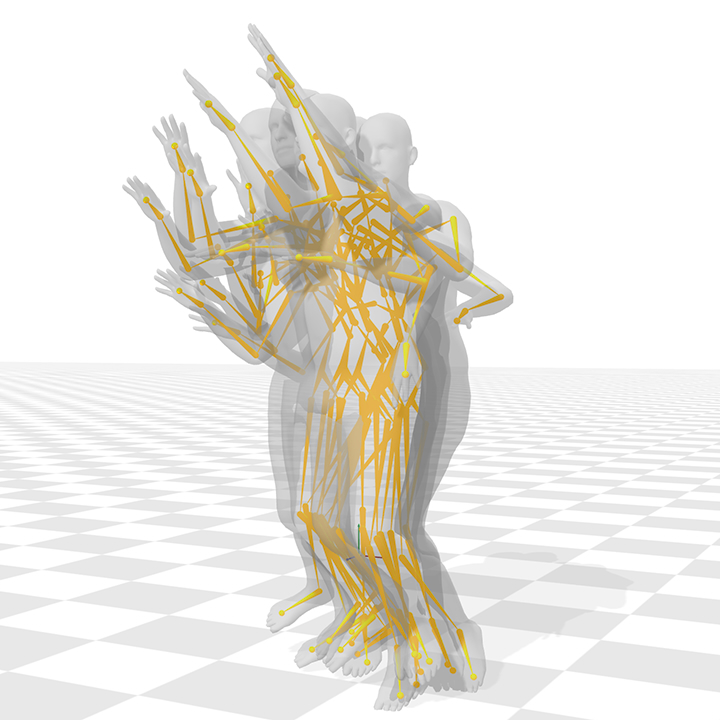}
        \caption{MotionAug(M2M).}
        \label{fig:motionaug_throw_demo}
    \end{subfigure}
    \begin{subfigure}{0.24\linewidth}
        \includegraphics[width=\linewidth]{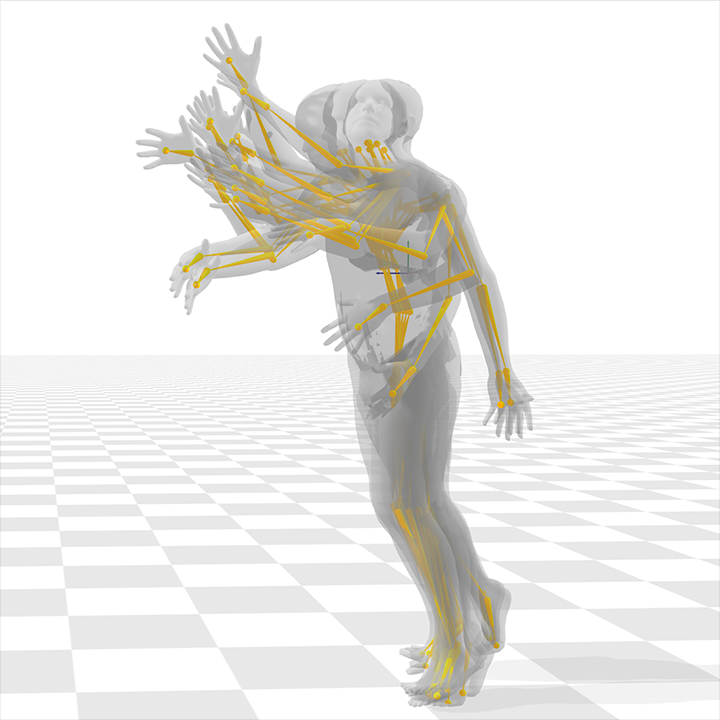}
        \caption{ACTOR(A2M).}
        \label{fig:actor_throw_demo}
    \end{subfigure}
    \begin{subfigure}{0.24\linewidth}
        \includegraphics[width=\linewidth]{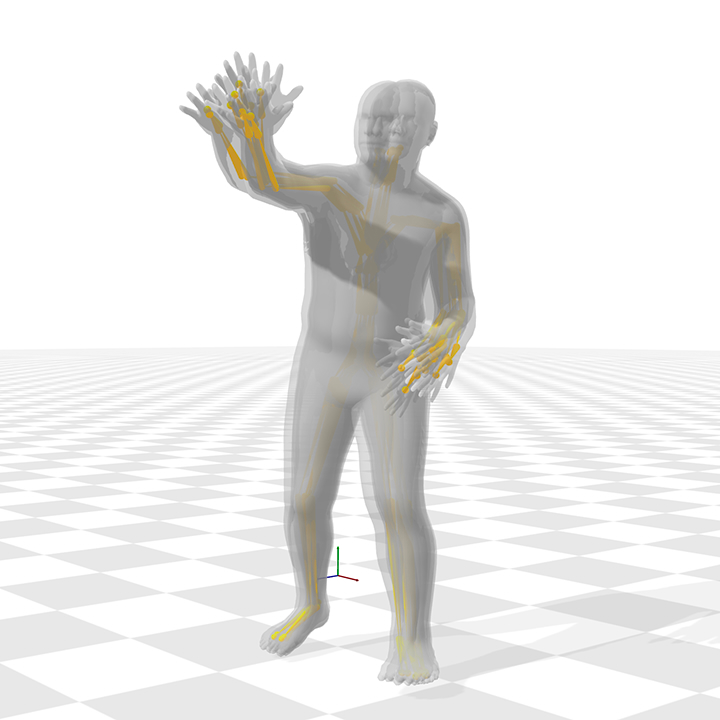}
        \caption{MDM-M2M.}
        \label{fig:mdm_m2m_throw_demo}
    \end{subfigure}
    \begin{subfigure}{0.24\linewidth}
        \includegraphics[width=\linewidth]{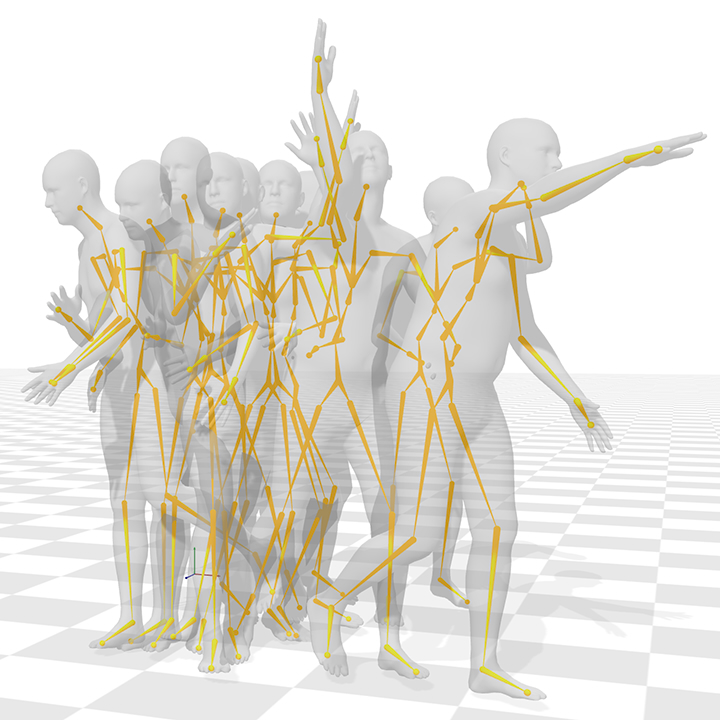}
        \caption{MDM-T2M.}
        \label{fig:mdm_throw_demo}
    \end{subfigure}
    \begin{subfigure}{0.48\linewidth}
        \includegraphics[width=\linewidth]{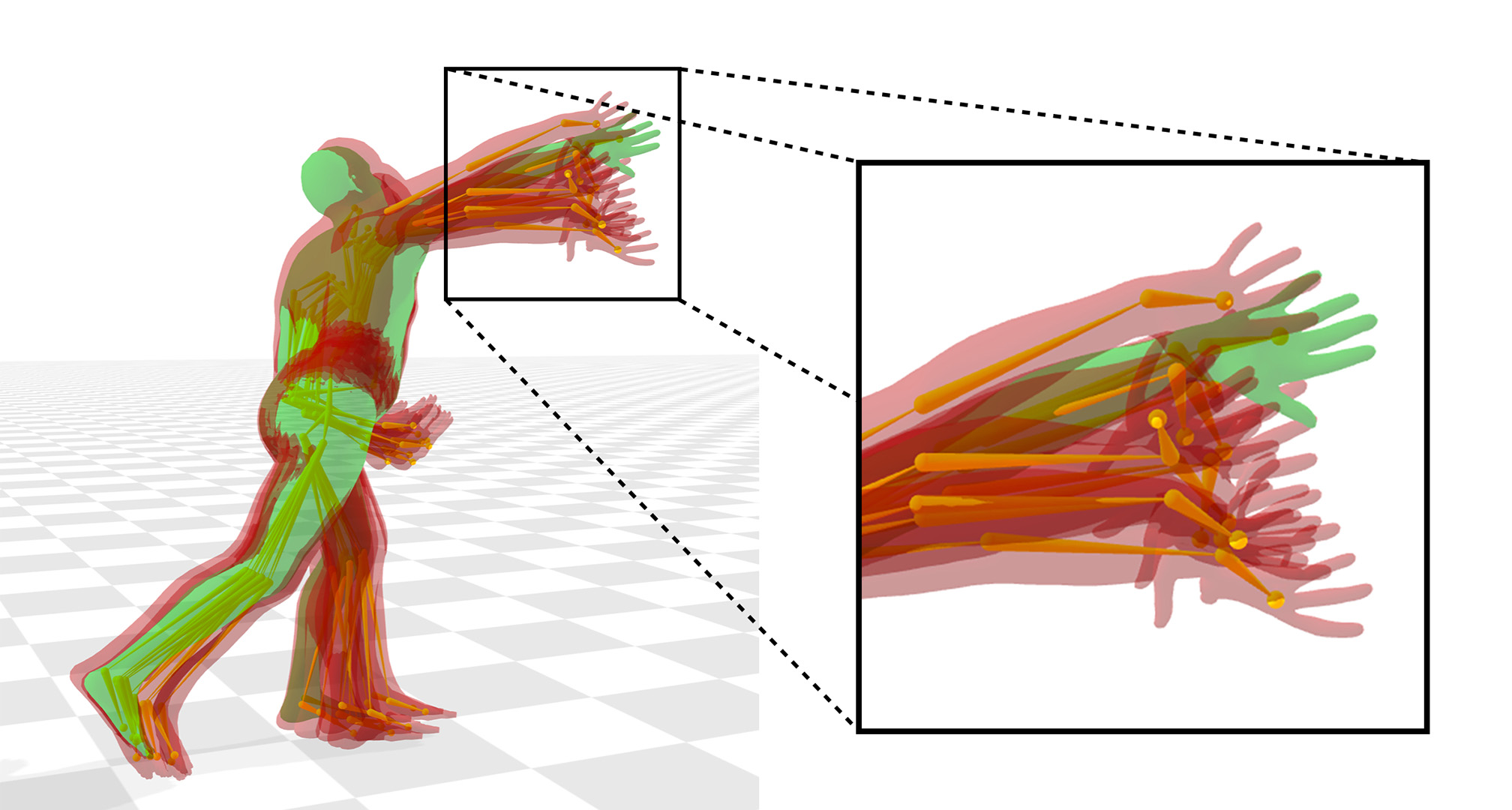}
        \caption{M2M pose generation by PoseAugment.
            The green pose is the ground truth, while the red ones are augmented.}
        \label{fig:augmented_pose}
    \end{subfigure}
    \begin{subfigure}{0.48\linewidth}
        \includegraphics[width=\linewidth]{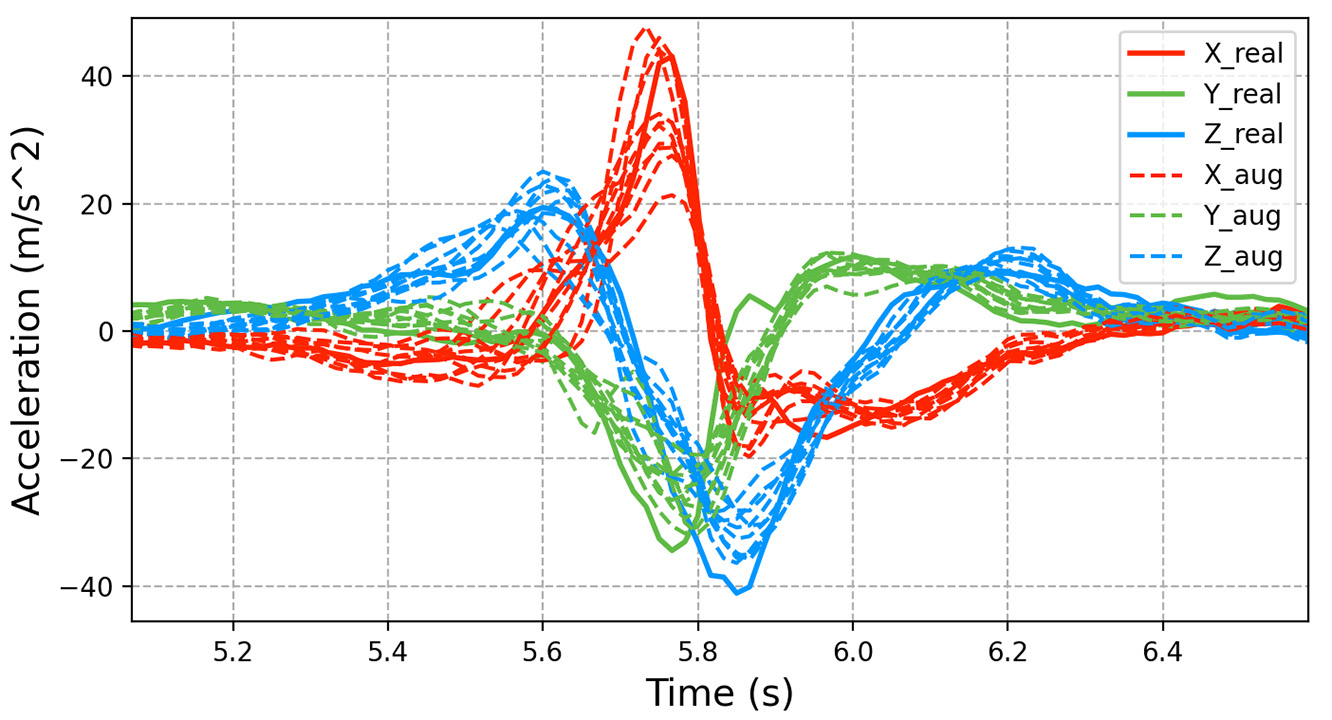}
        \caption{Augmented accelerations of the right wrist by PoseAugment.
            The solid lines are the ground truth, while the dotted lines are augmented.}
        \label{fig:augmented_imu}
    \end{subfigure}
    \caption{Visualization of the motion \textit{throwing a handball}.
        10 motion sequences are generated by MotionAug, ACTOR, MDM-M2M, MDM-T2M, and our method.}
    \label{fig:augmented_pose_and_imu}
\end{figure}

A huge burden in pose data collection is to
let the subjects repeat the same motion multiple times to cover the possible motion space.
PoseAugment simulates this process by generating poses that satisfy:
(1) \textbf{Fidelity}: the augmented poses should be close enough to the ground truth pose;
(2) \textbf{Diversity}: the deviations within augmented poses should be large enough
to cover the motion space.
Therefore, with PoseAugment, researchers can focus more on the diversity of motion types,
while reducing the motion repetition during data collection.

Qualitatively, we visualized the augmented poses by our baselines MotionAug, ACTOR,
MDM-M2M, and MDM-T2M (\cref{ssec:evaluation}), and our method respectively.
As shown in \cref{fig:augmented_pose_and_imu}, ACTOR and MDM-T2M are purely generated
from high-level action labels or texts.
The motion distribution is hard to control precisely, thus may violate
the original data distribution.
For other M2M methods, MotionAug uses a sequence-to-sequence model design,
thus the pose diversity is also hard to control at the frame level.
MDM-M2M is a variant of MDM, which denoises partially noised motions to generate
similar data, but suffers from motion diversity.
PoseAugment, on the other hand, generates poses frame-by-frame guided by the ground truth poses,
which best simulates the repetitions of the same motion.
The augmented IMU signals also achieve a high fidelity with the original data.
We found this M2M way performed the best in our experiments (\cref{ssec:evaluation}).
More qualitative results of our method are provided in Appendix B.

Quantitatively, we define $e_{pos}$ and $e_{rot}$, which are the positional
and rotational joint errors between ground truth pose and the corresponding
augmented poses, to reflect the reconstruction accuracy.
For diversity, we also defined $d_{pos}$ and $d_{rot}$, which are the mean positional
and rotational standard deviations within each joint of the augmented poses.
We augmented 4 times of new pose data on the AMASS dataset and got $e_{pos}=4.54cm,
d_{pos}=0.77cm, e_{rot}=10.86^{\circ}, d_{rot}=1.29^{\circ}$ over 24 joints.

\subsection{Physical Plausibility}\label{ssec:physical_plausibility}

\begin{figure}[ht]
    \centering
    \includegraphics[width=0.8\linewidth]{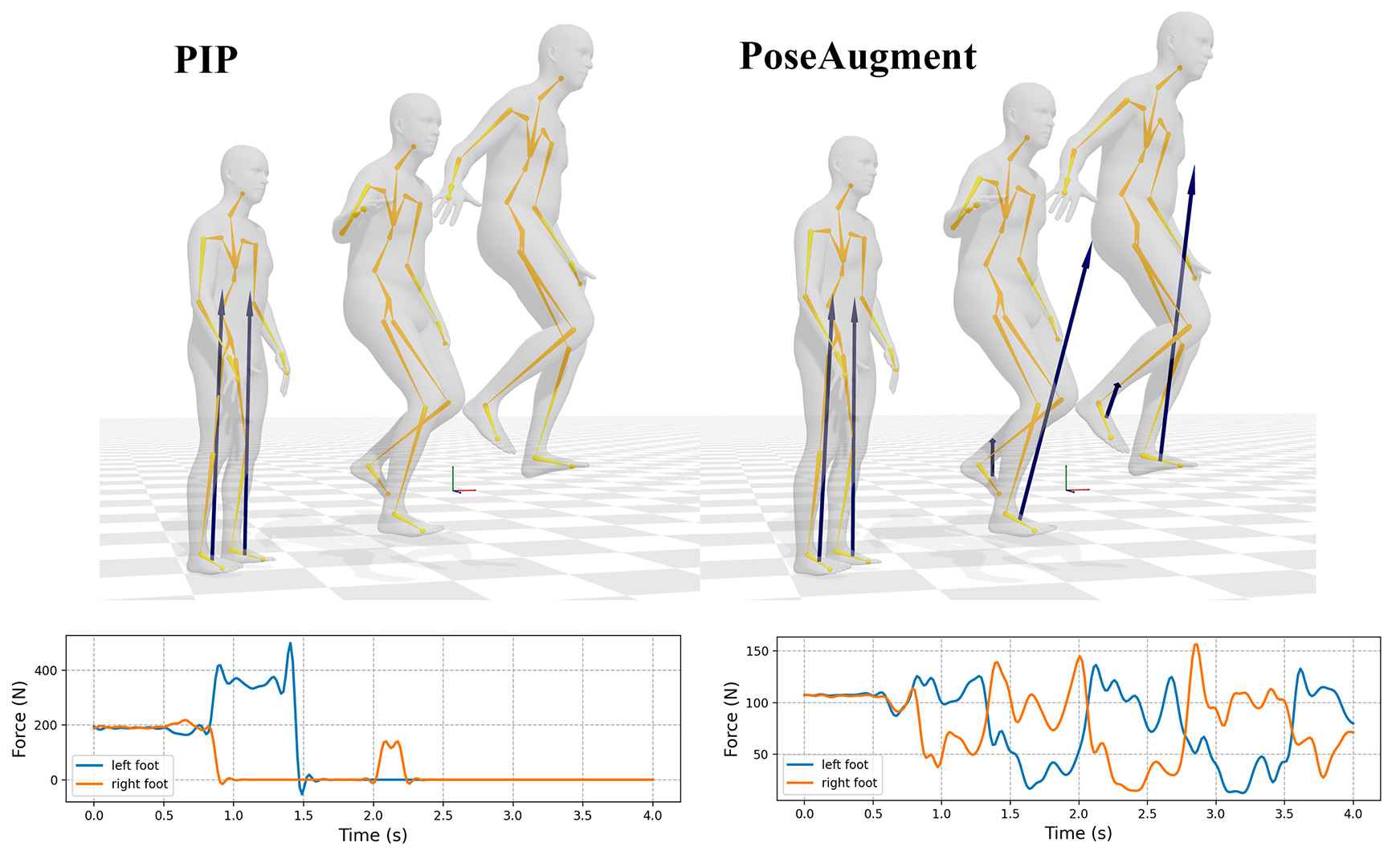}
    \caption{The reaction force estimation of climbing stairs.
        We visualized the force vectors on two feet by the blue arrows,
        and the time sequences of vertical reaction forces.
        PIP fails when the subject is off-ground,
        while our method does not have this limitation.}
    \label{fig:reaction_force}
\end{figure}

As discussed in \cref{sssec:dynamic_optimization}, we generalize the dynamic optimization
by discarding the ground plane assumption and the requirement of contact labeling.
We leverage the \textit{distance-to-root} principle and the \textit{stationary support}
assumption to achieve a soft estimation of the reaction forces,
broadening the supported motions compared with PIP\cite{Yi22}.

\cref{fig:reaction_force} demonstrates an example of climbing stairs.
The subject first stands at the bottom of the staircase and then climbs upward
along the steps using two feet alternatively.
Using previous threshold-based methods, the contact labels would be lost
when the subject is off-ground.
So, no reaction forces would be estimated.
PoseAugment optimizes poses purely based on dynamic properties without contact labels.
Though a small residual force may appear on unsuspended joints,
it achieves a softer and more pervasive pose dynamic estimation.

Furthermore, we quantitatively show that the physical optimization module also
improves the motion naturalness.
We measured the jitter $J$ (the 3rd derivative) of joint positions,
including the original data, poses generated by VAE without optimization,
and poses generated by VAE with optimization on the AMASS dataset.
We got $J_{ori}=2.11(100m/s^3)(SD=2.19)$, $J_{VAE}=5.15(SD=4.38)$, and $J_{ours}=2.32(SD=2.27)$.
These results show that the physical optimization significantly lowers the motion jitter
introduced by the VAE noises, bringing the motion naturalness to a similar level to the original data.

\subsection{Quantitative Evaluations}\label{ssec:evaluation}
\begin{table*}[ht]
    \centering
    \resizebox{\linewidth}{!}{
    \begin{tabular}{|c||c|c|c|c||c||c|c|c|c|}
        \hline
        Method & $e_{SIP}(^\circ)$ & $e_{rot}(^\circ)$ & $e_{pos}(cm)$ & $e_{mesh}(cm)$
        & Method & $e_{SIP}(^\circ)$ & $e_{rot}(^\circ)$ & $e_{pos}(cm)$ & $e_{mesh}(cm)$ \\
        \hhline{|=||====||=||====|}
        NoAug & $36.38$ & $17.12$ & $11.07$ & $12.42$ & NoAug & $26.20$ & $11.77$ & $7.40$ & $8.53$ \\
        \hhline{|-||----||-||----|}
        Jitter & $-2.2\%$ & $-3.0\%$ & $-4.9\%$ & $-4.2\%$ & Jitter & $-1.2\%$ & $0.2\%$ & $0.7\%$ & $0.3\%$ \\
        \hhline{|-||----||-||----|}
        MotionAug & $-5.4\%$ & $\mathbf{-13.5\%}$ & $-5.7\%$ & $\mathbf{-6.7\%}$ & MDM-M2M & $-0.5\%$ & $-0.7\%$ & $-0.1\%$ & $-0.2\%$ \\
        \hhline{|-||----||-||----|}
        Ours & $\mathbf{-14.5\%}$ & $-9.8\%$ & $\mathbf{-8.0\%}$ & $-5.4\%$ & Ours & $\mathbf{-7.6\%}$ & $\mathbf{-9.1\%}$ & $\mathbf{-7.8\%}$ & $\mathbf{-8.6\%}$ \\
        \hhline{|=||====||=||====|}
        NoAug & $30.64$ & $16.84$ & $8.62$ & $9.97$ & NoAug & $25.76$ & $11.77$ & $7.23$ & $8.37$ \\
        \hhline{|-||----||-||----|}
        Jitter & $-2.0\%$ & $-1.5\%$ & $+0.7\%$ & $+0.4\%$ & Jitter & $1.4\%$ & $-0.7\%$ & $+1.0\%$ & $+0.2\%$ \\
        \hhline{|-||----||-||----|}
        ACTOR & $-2.7\%$ & $-4.9\%$ & $-2.6\%$ & $-2.1\%$ & MDM-T2M & $-1.0\%$ & $-2.5\%$ & $+0.9\%$ & $+0.3\%$ \\
        \hhline{|-||----||-||----|}
        Ours & $\mathbf{-17.4\%}$ & $\mathbf{-20.7\%}$ & $\mathbf{-10.7\%}$ & $\mathbf{-12.2\%}$ & Ours & $\mathbf{-4.3\%}$ & $\mathbf{-8.3\%}$ & $\mathbf{-5.8\%}$ & $\mathbf{-6.8\%}$ \\
        \hline
    \end{tabular}}
    \caption{Comparisons of our method with Jitter, MotionAug, ACTOR, MDM-M2M, and MDM-T2M.
        Jitter and PoseAugment are tested on all datasets, while other methods are tested on their own dataset.
        The performance of the basic datasets are shown in absolute errors,
        while the performance of augmented datasets are shown in relative improvements
        compared with using the basic datasets.}
    \label{tab:result}
\end{table*}

To quantitatively evaluate PoseAugment compared with previous data augmentation methods,
we reproduced an IMU-based MoCap model TransPose\cite{Yi21}
and trained it with datasets augmented by different methods.
We use the performance of MoCap models to reflect the data augmentation quality.

\subsubsection{Baseline Methods.}

We compare PoseAugment with 5 other data augmentation methods:
(1)\textbf{Jitter}\cite{Yi21}: adding random noises $\sim \mathcal{N}(0, \sigma^2)$
directly to IMU data.
(2)\textbf{MotionAug}\cite{Maeda22}: a VAE/IK-based motion-to-motion augmentation method.
(3)\textbf{ACTOR}\cite{Petrovich21}: a VAE-based action-to-motion generative model.
(4)\textbf{MDM-T2M}\cite{Tevet22}: a diffusion-based text-to-motion generative model.
(5)\textbf{MDM-M2M}\cite{Tevet22}: a modification of the MDM model by denoising noised
    GT motions, such that it can augment data in the motion-to-motion manner like ours.

To our best knowledge, though noise-based data augmentation methods have been widely
adopted in classification tasks, they are rarely explored in IMU MoCap.
Recent generative models have also been applied to pose generation but have not been
evaluated in IMU-based MoCap either.
Therefore, we selected Jitter, MotionAug, ACTOR, and MDM as the most representative methods
to evaluate, covering the M2M, A2M, and T2M generation tasks.
The algorithm implementation details are shown in Appendix A.3.

\subsubsection{Datasets.}

Since MotionAug, ACTOR, and MDM are designed for different tasks (M2M, A2M, and T2M),
which need GT motions, action labels, and text descriptions respectively,
we followed the original papers and used them on different datasets
(HDM05\cite{Maeda22}, HumanAct12\cite{Guo20}, and HumanML3D\cite{Guo22}) respectively.
We also added MDM-M2M as a variant of MDM, since the diffusion model in MDM can also
generate motions in M2M manner.

To build the training datasets for the MoCap model, we first used these methods
to generate $1\times$ of their corresponding datasets, denoted as the \textit{basic datasets}.
They are treated as the unaugmented datasets in the following experiments.
Then, up to $4\times$ of motions are further generated using these methods,
denoted as the \textit{augmented datasets}.
For Jitter and PoseAugment, we directly applied them on the basic datasets to augment data.
We designed this setting for the reason that, the data distribution of baseline generative models
are largely different from their original training datasets.
We aim to compare the data quality of the generated data only, so the original data are discarded.

To simulate a common practice of using different amount of augmented data,
we use the basic datasets($1\times$), together with $1\times$ to $4\times$ of the augmented data,
to evaluate Jitter, MotionAug, ACTOR, MDM-M2M, MDM-T2M, and PoseAugment respectively.
To be consistent with previous practices\cite{Huang18,Yi21,Yi22}, we evaluate
all MoCap models on the real IMU data of the DIP test split.
The training details are described in Appendix A.4.

\subsubsection{Metrics.}

We evaluate MoCap models on:
(1)\textbf{SIP Error} $e_{SIP}$: the rotation error of upper arms and upper legs.
(2)\textbf{Rotation Error} $e_{rot}$: the rotation error of all body joints.
(3)\textbf{Position Error} $e_{pos}$: the position error of all body joints.
(4)\textbf{Mesh Error} $e_{mesh}$: the position error of all vertices on the SMPL mesh.
They are the same as the metrics used in\cite{Yi21,Yi22}.

\subsubsection{Results.}

\cref{tab:result} lists the data augmentation performance of PoseAugment compared with all baselines,
tested by training the TransPose model. 
The data augmentation performance is measured by the relative error reduction
compared to using the basic datasets only.
We found that Jitter has almost no impact on the model performance, while ACTOR, MDM-M2M, and MDM-T2M
are slightly better than Jitter, reducing the motion capture errors marginally.
PoseAugment outperformed the above three methods in all metrics,
showing a huge potential in improving the data quality for training MoCap models.
MotionAug reveals a comparable performance with PoseAugment.
However, as mentioned in \cref{ssec:pose_data_generation}, MotionAug needs users to annotate
IK keyframes, and is limited to 8 motion types, while PoseAugment does not have these limitations.

Notably, we further investigated how many training data are needed to achieve
the same level of model performance.
We trained TransPose on all the original AMASS\cite{Mahmood19} data (45.61h),
and got $e_{rot}=10.2^{\circ}$ and $e_{pos}=6.4cm$.
Using only the CMU dataset (a subdataset of AMASS, 9.06h) augmented with our method,
we achieved a comparable performance with $e_{rot}=10.2^{\circ}$ and $e_{pos}=6.5cm$,
but only using $19.9\%$ of training data.
It indicates a huge potential of PoseAugment to reduce the data collection burden in practice.

\subsubsection{Ablation Study.}


\begin{table}[ht]
    \centering
    \begin{tabular}{|c||c|c|c||c|c|c|}
        \hline
        \multirow{2}{*}{\makecell[c]{Method}} & \multicolumn{3}{c||}{$e_{rot}(^\circ)$} & \multicolumn{3}{c|}{$e_{pos}(cm)$} \\
        \cline{2-7}
        & MVAE+opt. & Ours-opt. & Ours & MVAE+opt. & Ours-opt. & Ours \\
        \hline
        MotionAug & $\mathbf{-13.6\%}$ & $-8.6\%$ & $-9.8\%$ & $\mathbf{-16.7\%}$ & $-4.7\%$ & $-8.0\%$ \\
        \hline
        ACTOR & $-20.3\%$ & $\mathbf{-21.7\%}$ & $-20.7\%$ & $-1.8\%$ & $\mathbf{-11.4\%}$ & $-10.7\%$ \\
        \hline
        MDM-M2M & $-8.7\%$ & $-8.0\%$ & $\mathbf{-9.1\%}$ & $-7.2\%$ & $-7.0\%$ & $\mathbf{-7.8\%}$ \\
        \hline
        MDM-T2M & $-7.3\%$ & $-8.1\%$ & $\mathbf{-8.3\%}$ & $-5.3\%$ & $-5.6\%$ & $\mathbf{-5.8\%}$ \\
        \hline
    \end{tabular}
    \caption{The ablation study on VAE model and physical optimization.
    Method indicates the datasets generated by the four techniques.
    Values are shown in the relative errors compared with using the basic datasets,
    the same as in \cref{tab:result}.}
    \label{tab:ablation}
\end{table}


We further performed a two-part ablation study on the VAE model and the physical optimization.
For the VAE part, our model has several network improvements over MVAE\cite{Ling20}
as introduced in \cref{ssec:pose_data_generation}.
We reproduced the original MVAE network, but kept the same motion frame representaion
and inferencing techniques, and the same physical optimization as our method,
denoted as \textit{MVAE+opt.} in \cref{tab:ablation}.
For the physical optimization part, we removed it from PoseAugment,
denoted as \textit{Ours-opt.} in \cref{tab:ablation}.
As a result, PoseAugment outperformed MVAE+opt. in ACTOR, MDM-M2M, and MDM-T2M datasets,
and outperformed Ours-opt. in MotionAug, MDM-M2M, and MDM-T2M datasets.

\section{Conclusion and Limitations}\label{sec:conclusion}

We propose PoseAugment, a novel human pose data augmentation method
that incorporates VAE-based pose generation and physical optimization.
Experiments demonstrate a significant improvement of PoseAugment
over previous pose augmentation methods, revealing a strong potential
of our method to alleviate the data collection burden in human pose-related tasks.

However, our method requires more computational cost compared with traditional methods.
Besides, the data augmentation performance is generally higher on smaller datasets.
It will benefit tasks with fewer available data or tasks
that involve personalization and few-shot learning the most.

\clearpage
\setcounter{page}{15}

\section*{Acknowledgements}

This work is supported by the Natural Science Foundation of China under Grant No. 62132010,
Beijing Key Lab of Networked Multimedia, Institute for Artificial Intelligence,
Tsinghua University (THUAI),
Beijing National Research Center for Information Science and Technology (BNRist),
2025 Key Technological Innovation Program of Ningbo City under Grant No.2022Z080,
Beijing Municipal Science and Technology Commission,
Administrative Commission of Zhongguancun Science Park No.Z221100006722018,
and Science and Technology Innovation Key R\&D Program of Chongqing.

%
%

\bibliographystyle{splncs04}
\bibliography{main}

\clearpage
\setcounter{page}{1}
\appendix

\section{Implementation Details}\label{sec:appendix_implementation_details}

We will provide the implementation details including the training
and evaluations of PoseAugment, for future researchers to reproduce our work.

\subsection{VAE Model Structure.}\label{ssec:appendix_vae_model_structure}

\begin{figure}[ht]
    \centering
    \includegraphics[width=0.5\linewidth]{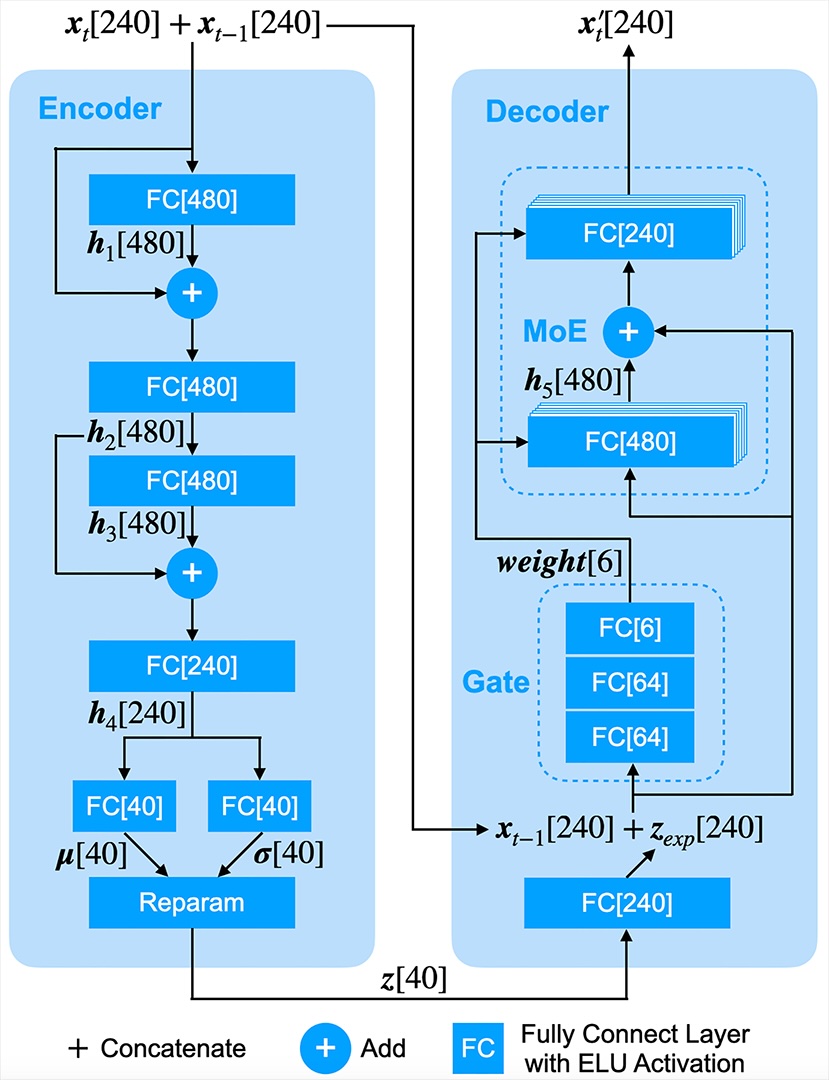}
    \caption{The VAE model structure details.
        Two adjacent frames are first input to the encoder with two separate residual blocks.
        After reparameterization, predictions of the current frame $\boldsymbol{x}_{t}'$ will be
        reconstructed by the decoder with the MoE architecture.}
    \label{fig:suppl_vae_structure}
\end{figure}

\cref{fig:suppl_vae_structure} demonstrates the structure details of our VAE model.
The current frame $\boldsymbol{x}_t$, together with the condition frame
$\boldsymbol{x}_{t-1}$, are first input to the encoder network to capture a latent
representation of their differences.
The encoder is comprised of 4 FC layers connected by two separate residual blocks.
Then, the encoding heads will output the mean and the standard deviation of
the latent vector, representing its distribution.
The reparameterization stage finally adds noises obeying the standard deviation
$\boldsymbol{\sigma}$ to the mean vector $\boldsymbol{\mu}$,
generating disturbed latent vectors.

Next, the latent vector is first decompressed to $\boldsymbol{z}_{exp}$, which will
be decoded together with the condition frame $\boldsymbol{x}_{t-1}$ by the decoder
with a MoE architecture\cite{Ling20}.
The MoE network consists of 6 identical expert networks.
Their output is smoothed by the weight from the gate network, generating a more stable
motion prediction.
Finally, the predicted frame $\boldsymbol{x}_{t}'$ will the current frame in the
reconstructed poses as well as the next condition frame in the next prediction.

In total, the VAE model contains $2950$k parameters, which is lightweight and easy to train.

\subsection{Training VAE Model.}\label{ssec:appendix_training_vae_model}

To make the autoregressive prediction stable, we adopted the \textit{scheduled sampling}
technique proposed by\cite{Ling20}.
First, the pose sequences are cut into mini-batches with lengths equal to 30.
Within each mini-batch, each pose prediction $\boldsymbol{x}_t'$ will be used
as the next conditioned frame $\boldsymbol{x}_{t-1}$ with probability $(1-p)$,
while the ground truth $\boldsymbol{x}_{t}$ will be used with probability $p$.
The total training epochs are divided into three stages, including
the supervised stage ($p=1$), the transition stage
($p$ decreasing from $1$ to $0$ linearly), and the autoregressive stage
($p=0$).
Since the VAE model during inferencing is purely autoregressive, this design
will make our model robust to self-prediction errors.
In practice, since the sampling rate is relatively high (60Hz, which means
the frame difference is small for each prediction), we choose $L=30$ (0.5s)
and the lengths of the three stages to be 50, 150, and 200 epochs.

In total, we trained the VAE model first with $10$ warm-up epochs, where the
learning rate increased from $2\times10^{-6}$ to $2\times10^{-5}$ linearly.
Then, the model was trained with $400$ scheduled sampling epochs in total,
where the learning rate started from $2\times10^{-5}$ and decayed
exponentially with a factor of $0.99$ for each epoch.
We used the Adam optimizer, with a batch size equal to $512$.

\subsection{Baseline Methods}\label{ssec:appendix_baseline_methods}

Our baselines include Jitter, MotionAug\cite{Maeda22}, ACTOR\cite{Petrovich21},
and MDM\cite{Tevet22} (including MDM-M2M and MDM-T2M).
Jitter is a universal data augmentation method, that could be applied to any data modalities.
MotionAug is a VAE/IK-based method, trained on HDM05, and only supports 8 motion types.
ACTOR and MDM are conditioned on action labels and text descriptions,
which also require specific annotated datasets.
Therefore, we first applied these methods on their corresponding datasets
to generate the basic datasets, as described in \cref{ssec:evaluation}.
After that, we applied these methods again, together with Jitter and PoseAugment
to generate the augmented datasets.

For Jitter, we add random noise $\sim N(0, \sigma^2)$ to the IMU data, following\cite{Yi21,Yi22}.
To find the best $\sigma$, we first conducted a pilot study, using different $\sigma$
to train the TransPose model, and selected the best $\sigma$ with minimal reconstruction errors.
We searched $\sigma$ within $\{1\times10^{x}, 2\times10^{x}, 5\times10^{x}\}, x\in\{-1,-2,-3,-4\}$,
and got the best $\sigma=0.002$.

For MotionAug, since the IK-based method needs users to annotate the motion keyframes
(referred to as "semi-automatic" in the original paper), we directly used 1/5 of the
released dataset as the basic dataset, and 4/5 as the augmented dataset.

For ACTOR, we generated 200 motion clips for each action (12 actions in total)
as the basic dataset, and 800 motions for each action as the augmented dataset.
Each motion clip is sampled at 20Hz and contains 70 frames.
Then, all motion clips are upsampled to 60Hz using quadratic interpolation,
to be consistent with TransPose.

For MDM-T2M, we generate 1 motion clip for each text in the HumanML3D text split
(4384 texts in total) as the basic dataset, and 4 motions for each text as the augmented dataset.
Each motion clip is sampled at 20Hz and contains 120 frames.
We upsampled the motions to 60Hz as well to train the TransPose model.

For MDM-M2M, we modified the diffusion model in the original MDM, such that it can denoise
partially noised motions.
We chose the partial noising steps to be 500 (half of the original MDM model),
such that the generated motion diversity
($d_{pos}=0.71cm, d_{rot}=2.42^\circ$, the same metric as in \cref{ssec:fidelity_and_diversity})
is comparable with PoseAugment ($d_{pos}=0.82cm, d_{rot}=1.91^\circ$).
Then, we generated the basic and augmented datasets using the modified MDM-M2M model
in a similar way as MDM-T2M.

To generate the augmented datasets for Jitter and PoseAugment, we used them on all of
the basic datasets generated by MotionAug, ACTOR, MDM-M2M, and MDM-T2M.
As last, we got the basic and augmented datasets for all baselines and PoseAugment
to evaluate their data augmentation performance.

\subsection{MoCap Model Training}\label{ssec:appendix_mocap_model_training}

To evaluate data augmentation methods for training MoCap models,
two key factors need to be addressed:
(1) How much basic data should be used?
(2) How much data should be augmented?
They both affect the actual dataset size to train the model,
which is essential for the data augmentation performance,
as explored in\cite{Zheng23}.

For the basic data, the lengths of the datasets generated by MotionAug,
ACTOR, and MDM are 2.65h, 2.22h, and 8.89h respectively, which are chosen to be
comparable with the training dataset size of these methods.
They simulate the situation of using small or big datasets in real practice.
For the augmented data, we define the augmentation scale $n_{aug}$,
which represents using $1\times$ of the basic dataset together with $(n_{aug}-1)\times$
of the augmented dataset to train the model.
We found the data augmentation performance would converge quickly
when $n_{aug}$ reaches about $5$ ($4\times$ of augmented data),
and more data would not improve the model performance.
Therefore, we set $n_{aug}$ to be $2-5$ in our evaluation
and selected the best model in each training to simulate
the tuning process on $n_{aug}$.

The training data are all resampled to 60Hz and are first cut to a fixed window size
of 200 before training.
For the basic dataset generated by ACTOR (2.22h),
we trained the TransPose model for $200$ epochs with batch size $64$.
When using other datasets or a different $n_{aug}$, we modified the training epochs
accordingly to make sure the total training steps would be the same.
We used the Adam optimizer, with the learning rate decreasing linearly from
$5\times10^{-4}$ to $5\times10^{-5}$.

\section{Qualitative Results}\label{sec:appendix_qualitative_results}

Here we further provide more motions generated by PoseAugment,
to demonstrate the generalizability of our method.

\begin{figure}[ht]
    \centering
    \begin{subfigure}{0.24\linewidth}
        \includegraphics[width=\linewidth]{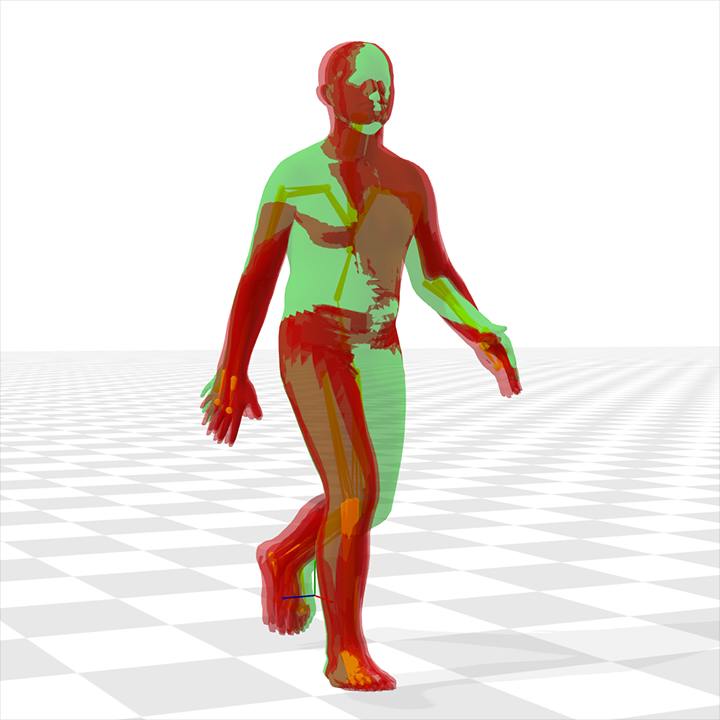}
        \caption{Normal walking.}
    \end{subfigure}
    \begin{subfigure}{0.24\linewidth}
        \includegraphics[width=\linewidth]{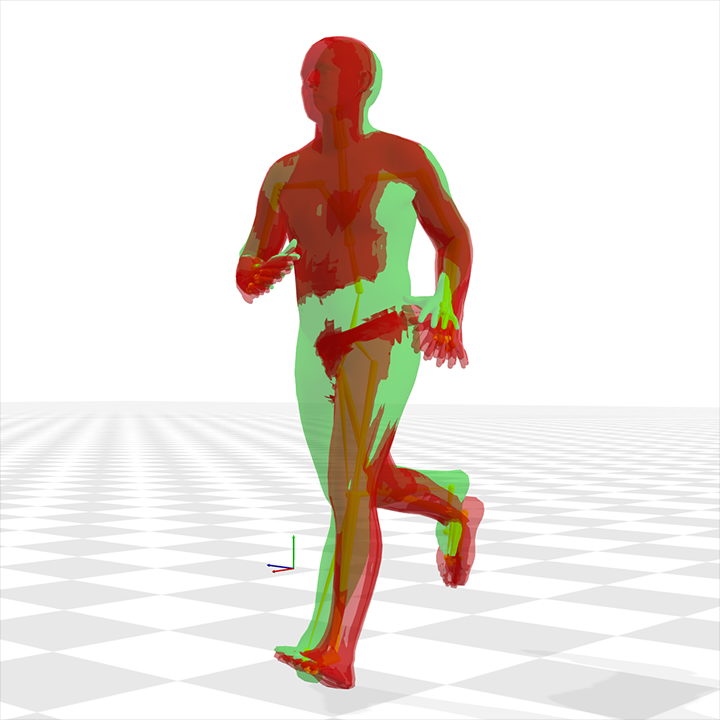}
        \caption{Running.}
    \end{subfigure}
    \begin{subfigure}{0.24\linewidth}
        \includegraphics[width=\linewidth]{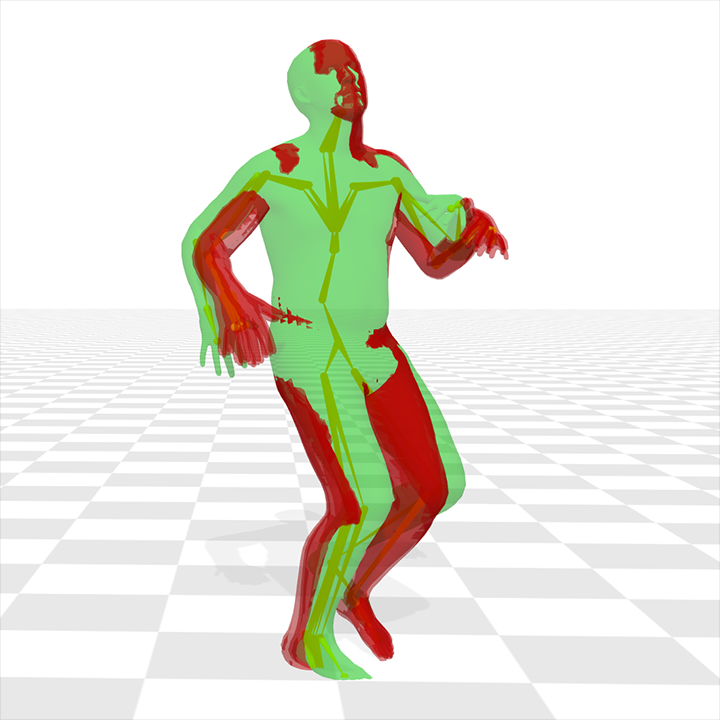}
        \caption{Running backward.}
    \end{subfigure}
    \begin{subfigure}{0.24\linewidth}
        \includegraphics[width=\linewidth]{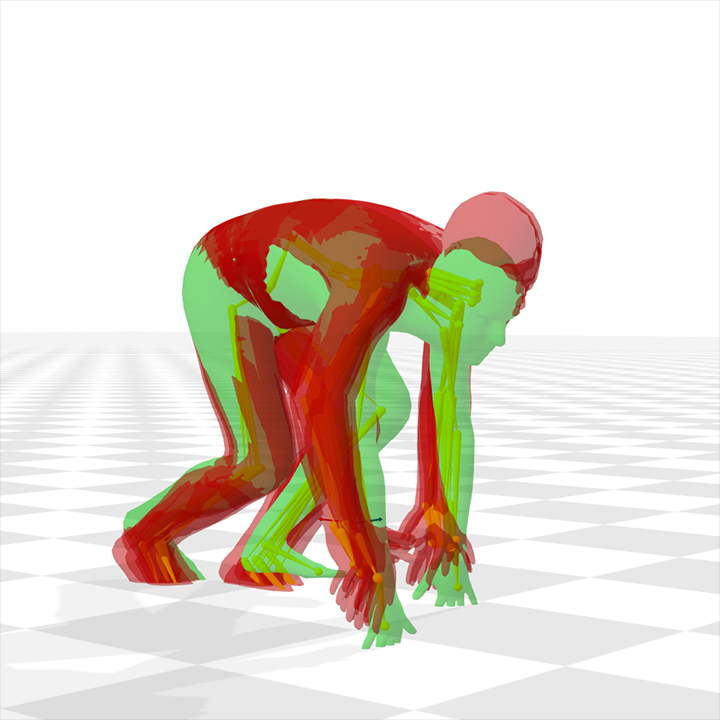}
        \caption{Crouching.}
    \end{subfigure}
    \begin{subfigure}{0.24\linewidth}
        \includegraphics[width=\linewidth]{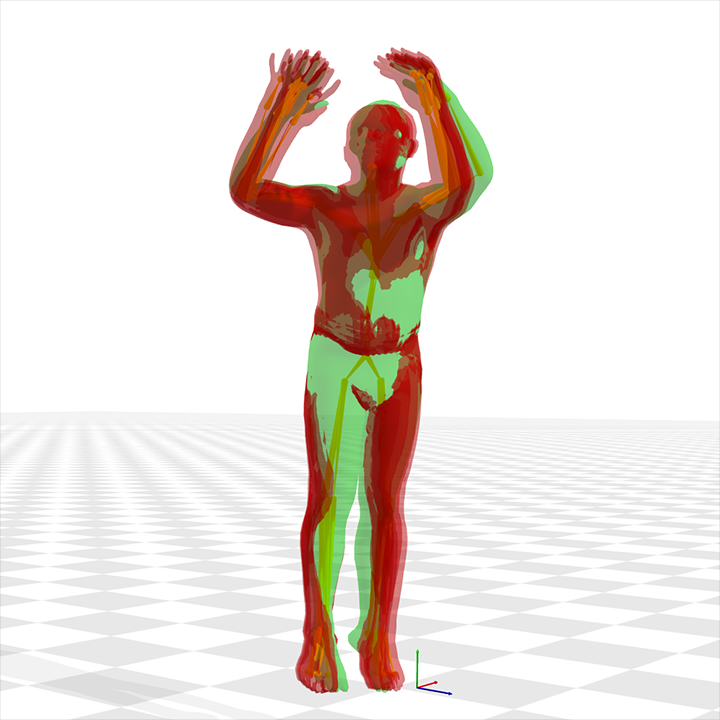}
        \caption{Jumping high.}
    \end{subfigure}
    \begin{subfigure}{0.24\linewidth}
        \includegraphics[width=\linewidth]{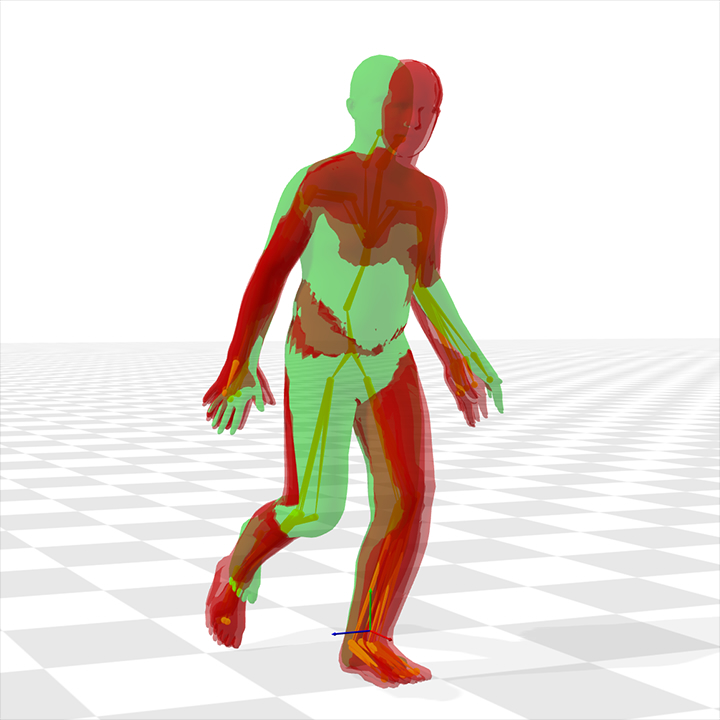}
        \caption{Jumping on 1 foot.}
    \end{subfigure}
    \begin{subfigure}{0.24\linewidth}
        \includegraphics[width=\linewidth]{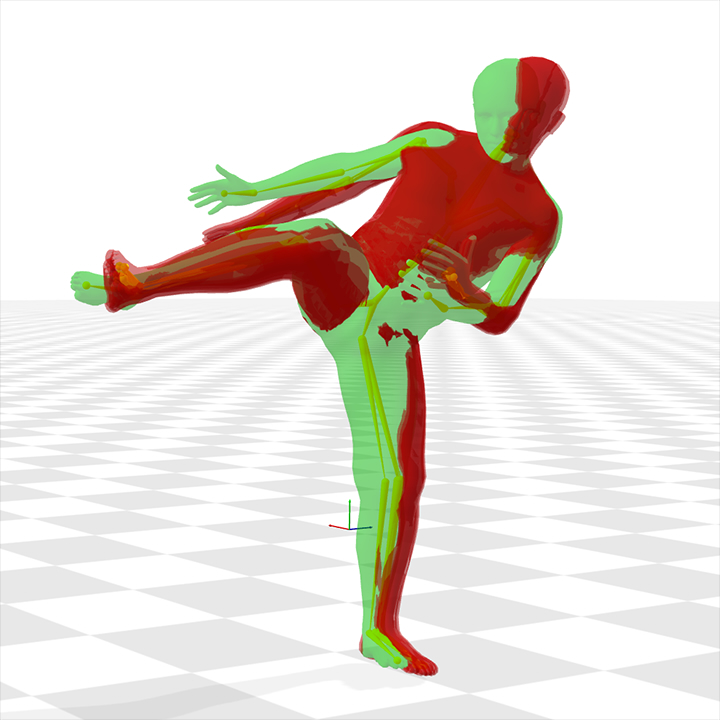}
        \caption{Kicking.}
    \end{subfigure}
    \begin{subfigure}{0.24\linewidth}
        \includegraphics[width=\linewidth]{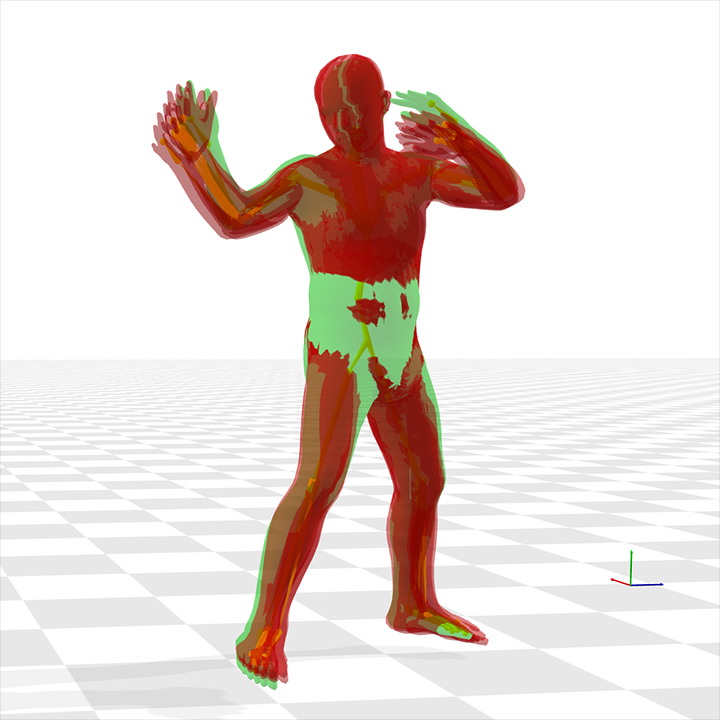}
        \caption{Punching.}
    \end{subfigure}
    \begin{subfigure}{0.24\linewidth}
        \includegraphics[width=\linewidth]{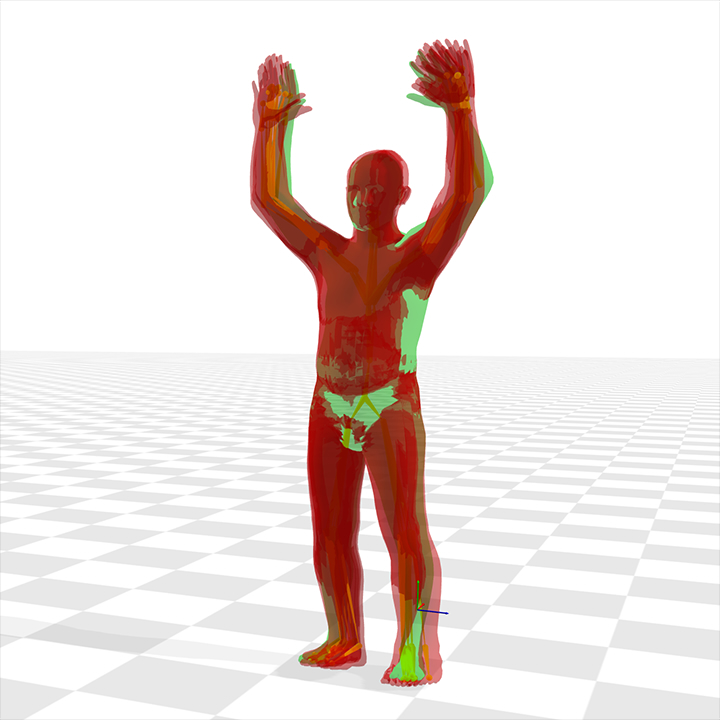}
        \caption{Waving hands.}
    \end{subfigure}
    \begin{subfigure}{0.24\linewidth}
        \includegraphics[width=\linewidth]{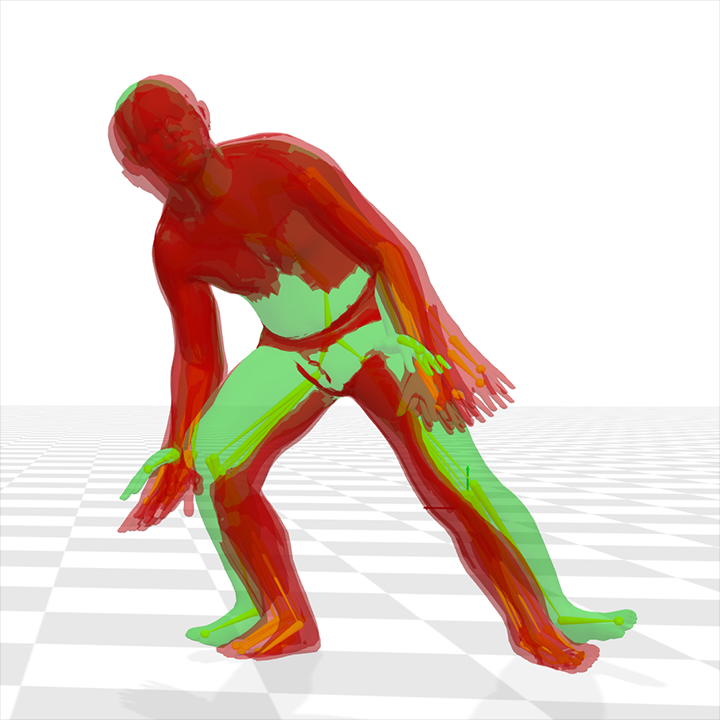}
        \caption{Dodging.}
    \end{subfigure}
    \begin{subfigure}{0.24\linewidth}
        \includegraphics[width=\linewidth]{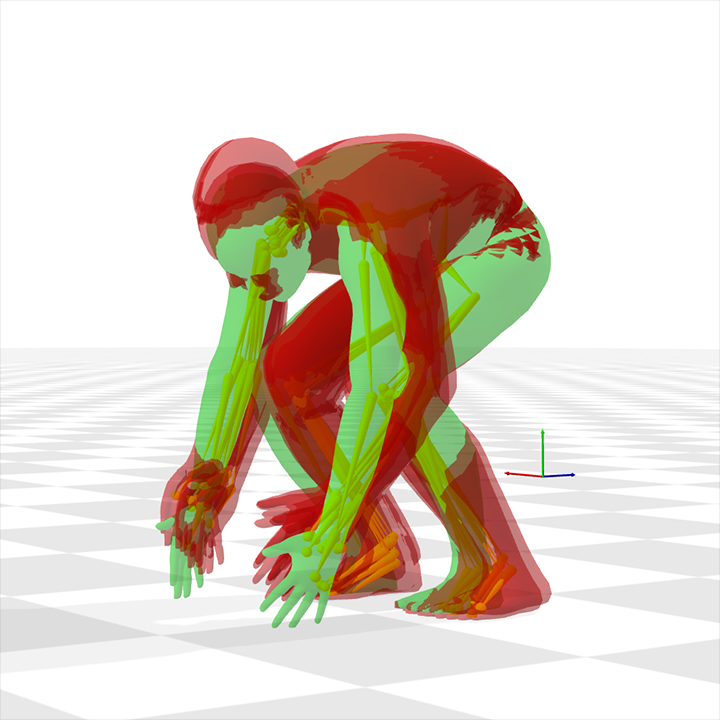}
        \caption{Bending down.}
    \end{subfigure}
    \begin{subfigure}{0.24\linewidth}
        \includegraphics[width=\linewidth]{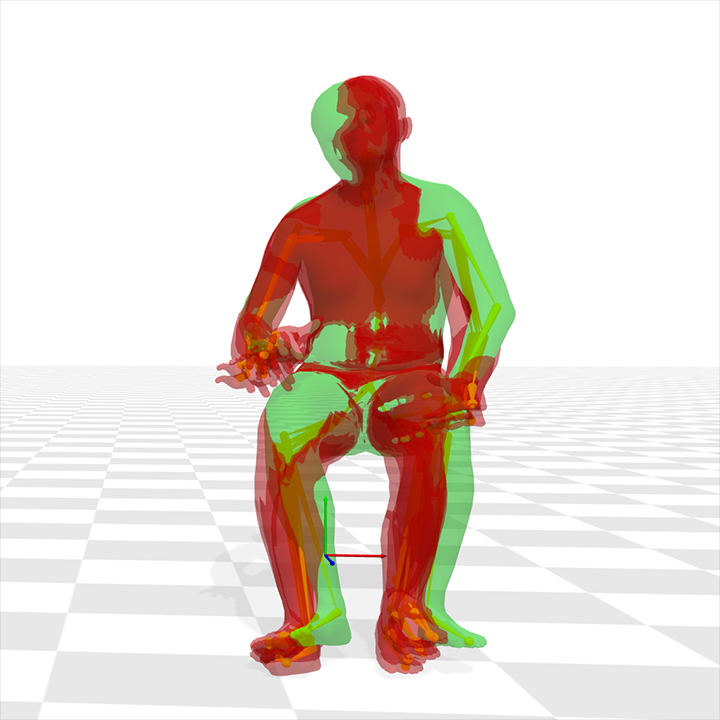}
        \caption{Sitting down.}
    \end{subfigure}
    \caption{More poses generated by PoseAugment, including various motion types.
        In each subfigure, one ground truth pose (green) and 9 augmented poses (red) are visualized.}
    \label{fig:appendix_qualitative_results}
\end{figure}

As shown in \cref{fig:appendix_qualitative_results},
we randomly selected some motions with different motion types from the AMASS dataset,
and augmented 9 similar motions with PoseAugment for each of them.
As a result, the augmented poses followed the original motions closely,
but with more diversity to cover the motion space.
It simulates the repetitions during data collection.
Compared with other M2M, A2M, and T2M methods, our method is not limited to specific motion types,
and maintains the original data distribution well, which is essential for the data augmentation task.

\section{Discussion}\label{sec:appendix_discussion}

We would like to discuss the key factors of our data augmentation method,
the comparison with diffusion-based methods, and the potential applications of our method,
which will provide more insights into our design choices,
and benefit future research that has similar goals to us.

\subsection{Data Augmentation Performance}\label{ssec:appendix_data_augmentation_performance}

Fundamentally, all data-driven tasks expect the test data (in real use) to
have a similar distribution with the training data, so that the knowledge the model
acquires during training can be transferred to the test data.
As a result, to achieve the best test performance, we need the training data
distribution to be closer to the test data distribution, and the training data
should cover this distribution comprehensively.

In our work, we found that the data augmentation performance is related to many factors,
which confirms the above analysis.
The first factor is the dataset size.
With a fixed training time, models trained on the HumanML3D (8.89h) have a better accuracy
compared with models trained with HumanAct12 (2.22h).
Models trained with the augmented dataset would also outperform using the original dataset
(\cref{ssec:evaluation}).
We also found the data augmentation performance of PoseAugment is generally higher on
smaller datasets, due to the problem of overfitting.
Thus, our method would benefit the tasks with high data collection costs or involve few-shot learning the most.

Another factor is the data distribution.
We found direct manipulation of the IMU data (\eg Jitter) would not improve the model performance much,
since it may fail to capture the physical constraints of the body joints, thus lowering the data quality.
Our method, on the other hand, synthesizes IMU signals from augmented poses with physical plausibility.
This would best fit the distribution of the real MoCap data.

Last but not least, the data diversity is also important.
From the experiments, we conclude that the best structure of the training dataset is "diverse motion types
with proper repetitions", just like the normal data collection process.
Since the ACTOR and MDM-T2M models are only conditioned on high-level information,
the poses they generate have too much diversity and lack of repetitions.
It would make the model hard to converge, resulting in underfitting.
PoseAugment only generates poses with a high fidelity, while with appropriate diversity to simulate
the motion repetitions during data collection.
This would make the model easier to converge on each motion, but not overfit to a specific motion
pattern compared with no data augmentation at all.
As a result, it achieved the best performance in our experiments.

\subsection{Compare with Diffusion-based Methods}\label{ssec:appendix_compare_with_diffusion}

Since diffusion models have been widely used in AIGC and human motion diffusion models
\cite{Zhang24,Tevet22,Chen23,Yuan23,Karunratanakul23}
have also been proposed to generate poses, we would like to discuss why we chose VAE instead of
diffusion models in our work.

Our first priority is data fidelity.
Current human diffusion models are good at generating high-quality data from random noises
and condition information, but not focus on reconstructing poses.
The forward and reverse diffusion process may introduce large fluctuations to the pose distribution.
Our method reconstructs motion frames with minimal errors to ensure the frame level
consistency for synthesizing IMU data.
Therefore, we choose VAE over diffusion models to achieve a higher data fidelity.

The second factor is data generation efficiency.
Diffusion models are known for a longer inferencing time\cite{Chen23} due to the iterative
diffusion process, thus taking a significantly longer time to generate data.
Moreover, another practical reason is that current human diffusion models are trained on
the HumanML3D dataset, which only generates global joint positions.
They need first to be converted into local joint rotations using Inverse Kinematics (IK),
which are even more time-consuming than generating joint positions.
As a result, our method is $33\times$ faster than MDM\cite{Tevet22}
and is a more pervasive data augmentation method.

\subsection{Applications}\label{ssec:appendix_applications}

PoseAugment aims to improve the model performance and alleviate
the data collection burden for IMU-based motion capture.
Furthermore, since the human pose is a general representation of human motion,
our method can be used for any tasks driven by poses but is not limited to augmenting IMU data.
For example, PoseAugment can directly benefit pose-based action recognition,
anomaly detection, and motion rendering tasks.
The augmented poses can also be converted to other modalities, like images or videos,
to benefit CV-based motion capture and recognition.
The physical module estimates the dynamic properties of the human body,
which can be adopted for motion-related diseases and sports analysis.
PoseAugment can also make contributions to early prototyping and explorations in research
when there are few available data to use before the mass data collection.
In a word, as long as data-driven approaches continue to be widely employed,
PoseAugment will bring value to the aforementioned domains.

\end{document}